\documentclass[lettersize,journal]{IEEEtran}
\usepackage{amsmath,amsfonts}
\usepackage{algorithmic}
\usepackage{algorithm}
\usepackage{array}
\usepackage[caption=false,font=normalsize,labelfont=sf,textfont=sf]{subfig}
\usepackage{textcomp}
\usepackage{stfloats}
\usepackage{url}
\usepackage{verbatim}
\usepackage{graphicx}
\usepackage{cite}
\usepackage{float}
\usepackage{placeins}
\usepackage{lipsum}
\usepackage{soul,color}
\usepackage{subcaption}
\usepackage{pdflscape}
\usepackage{rotating}
\usepackage{makecell}
\usepackage{placeins}
\usepackage{booktabs}
\usepackage{caption}

\begin{document}

\title{Communication-Aware Reinforcement Learning for Cooperative Adaptive Cruise Control}

\author{Sicong Jiang*, Seongjin Choi*, and Lijun Sun~\IEEEmembership{Senior Member,~IEEE,}
        
\thanks{*Co-first authors: S. Jiang and S. Choi contributed equally to this paper. Corresponding author: Seongjin Choi.}
\thanks{S. Jiang and L. Sun are with the Department of Civil Engineering, McGill University, Montreal, Canada. (e-mail: sicong.jiang@mail.mcgill.ca; lijun.sun@mcgill.ca).}
\thanks{S. Choi is with the Department of Civil, Environmental, and Geo- Engineering, University of Minnesota, Minneapolis, USA. (e-mail: chois@umn.edu)}
}



\maketitle

\begin{abstract}
Cooperative Adaptive Cruise Control (CACC) is essential for enhancing traffic efficiency and safety in Connected and Autonomous Vehicles (CAVs). Reinforcement Learning (RL) has proven effective in optimizing complex decision-making processes in CACC, leading to improved system performance and adaptability. Among RL approaches, Multi-Agent Reinforcement Learning (MARL) has shown remarkable potential by enabling coordinated actions among multiple CAVs through Centralized Training with Decentralized Execution (CTDE). However, MARL often faces scalability issues, particularly when CACC vehicles suddenly join or leave the platoon, resulting in performance degradation. To address these challenges, we propose Communication-Aware Reinforcement Learning (CA-RL). CA-RL includes a communication-aware module that extracts and compresses vehicle communication information through forward and backward information transmission modules. This enables efficient cyclic information propagation within the CACC traffic flow, ensuring policy consistency and mitigating the scalability problems of MARL in CACC. Experimental results demonstrate that CA-RL significantly outperforms baseline methods in various traffic scenarios, achieving superior scalability, robustness, and overall system performance while maintaining reliable performance despite changes in the number of participating vehicles.
\end{abstract}



\section{Introduction}
Autonomous Vehicle (AV) technology has been studied extensively in recent years, which is expected to provide a safe and efficient transportation system in the future. An ideal AV would be capable of driving without any human intervention by utilizing sub-systems such as the perception system and driving logic system \cite{campbell2010autonomous, azad2019fully,choi4161343framework}. An important basic function of autonomous vehicles is speed control, enabling vehicles to autonomously adjust their speed and distance to maintain a safe gap from the vehicle ahead\cite{schwarting2018planning,faisal2019understanding, tak2022safety}. Therefore, speed control models for autonomous vehicles have become increasingly popular in recent years\cite{xu2018accurate}.

\textit{Cooperative Adaptive Cruise Control (CACC)} is a widely studied vehicle control method for connected vehicles. Its early foundations are \textit{Cruise Control (CC)}, which only controls the vehicle to drive at a specific speed, and \textit{Adaptive Cruise Control (ACC)} \cite{doi1994development,fujita1995radar,seiler1998development}, which controls speed according to the proceeding vehicle's information. ACC is designed to maintain a specific distance behind a preceding vehicle and is believed to improve roadway capacity, safety, and fuel efficiency \cite{vahidi2003research,dey2015review,do2019simulation,lee2019development}. However, recent studies have suggested that the positive effects of ACC on the roadway were not fulfilled by currently available ACC-equipped vehicles from various manufacturers such as Tesla, Mercedes-Benz, and BMW \cite{ciuffo2021requiem}. Therefore, CACC, which combines automated speed control with the vehicle communication system, highlights its necessity. There are usually two communication topologies in CACC, Vehicle-to-Vehicle (V2V) and Vehicle-to-Infrastructure (V2I) communication \cite{shladover2015cooperative}. V2V uses communication between vehicles to bring direct information exchange, while V2I connects vehicles and Infrastructure to bring more comprehensive information integration. Each communication topology has its own advantages. Different methods are often selected based on different task objectives in applications of CACC.


In addition to communication systems, the vehicles' control model of CACC is also a famous topic of research in academia. Many feedback control models based on control theory are widely used such as Proportional-Integral-Derivative (PID) Control \cite{zhu2023research}, Fuzzy Logic Control \cite{maged2023optimal}, and Model Predictive Control (MPC) \cite{kural2010model}. 
%
%
These feedback control methods calculate precise distance and speed adjustments by considering the difference between actual control output and ideal output and combining vehicle communication information.
However, they also face challenges, such as the complexity and computational burden of advanced control methods like MPC \cite{yu2013model}. Also, it's difficult for them to handle the nonlinear nature of CACC systems \cite{manenti2011considerations, fiedler2023mpc}.

In recent years, novel machine learning methods such as Reinforcement Learning (RL) have gained attention in ACC/CACC-related research as a promising complementary method to traditional control models. RL is used to describe and solve the problems of maximizing rewards or realizing specific goals through sequential decision-making adopted by the agents \cite{kaelbling1996reinforcement, arulkumaran2017deep}. In ACC/CACC, vehicles can learn longitudinal control strategies based on the designed rewards function. After proper training and tuning, vehicles can achieve safe, efficient, and stable longitudinal control performance \cite{zhu2020safe}. RL can handle nonlinear models well, and its performance can be improved as continuous data input, which makes it very suitable for CACC.

In RL-based CACC research, there are two main key aspects to consider when designing the model: 1) policy consistency and 2) full use of traffic flow information. First, policy consistency refers to whether the RL model can ensure that each vehicle (agent) in CACC has the same (or a similar) policy. Policy consistency can guarantee that the RL model for CACC has good scalability in terms of the number of vehicles in the platoon. Second, the full use of traffic flow information refers to whether the vehicle can obtain and use the entire traffic flow information before making control actions. This can ensure that the RL model can output optimized control actions which can improve the entire traffic flow.
However, the policy consistency and the full use of traffic flow information are not easy to obtain both at the same time. 
For example, Single Agent Reinforcement Learning (SARL) \cite{desjardins2011cooperative} uses data from in-vehicle sensors and local communication data (usually from only one preceding vehicle) to train individual vehicles, which achieves policy consistency but often lacks information about the entire traffic flow, hampering its effectiveness in the larger platoon. 
Conversely, Multi-Agent Reinforcement Learning (MARL) \cite{chu2019model, chu2020multi, peake2020multi} collects information through a centralized information center and trains the learning of each agent, which allows the algorithm to make full use of traffic flow information. However, since MARL usually assigns a different policy for each agent, it faces challenges with policy consistency across different vehicles, especially as the number of vehicles changes. This inconsistency of policy can affect system reliability and safety when there are more vehicles in the platoon than in the trained scenario. 
The goal of RL-based CACC is to develop a model that can utilize the information of the entire traffic flow while allowing each vehicle to use a relatively consistent policy to ensure stability and scalability. 

 Therefore, we developed a novel framework known as Communication-Aware Reinforcement Learning (CARL). This framework skillfully merges the strengths of Single-Agent Reinforcement Learning (SARL) and Multi-Agent Reinforcement Learning (MARL), which allows our model to not only maintain policy consistency but also effectively gather and utilize traffic flow data.
Our approach ensures that the CACC model remains highly scalable, optimizing the usage of traffic flow information for collaborative control purposes. Furthermore, CARL is designed for seamless integration with existing models, thus offering a flexible and valuable upgrade to current RL models employed in a variety of CACC scenarios.

In summary, the contribution of our proposed work is :

\begin{itemize}

\item 
We developed the Communication Aware Reinforcement Learning (CARL) framework, and redesigned the communication architecture based on V2V, thereby enhancing the adaptability and efficiency of RL in complex traffic environments.

\item 
We introduced a flexible inter-vehicle information transfer mechanism within CARL, compatible with various RL algorithms, enabling broader application across different CACC systems.

\item
By merging the strengths of SARL and MARL, our algorithm can take into account both policy consistency and traffic flow information, which significantly improved CARL's generalization ability, ensuring stable performance in a variety of traffic scenarios.

\end{itemize}

The rest of the paper is organized as follows. Section \ref{sec:methodology} describes the detailed methodology and setup of our model. Section \ref{sec:expset} describes our experimental setting and setup. Section \ref{sec:result} presents the results and analysis of the experiments. Finally, in Section \ref{sec:conclusion}, we present summary of this paper with contributions and limitations of this paper, as well as future research directions.

\section{Methodology}\label{sec:methodology}

\subsection{Problem Formulation}
Here, we formulate the problem of interest as a Markov Decision Process (MDP). An MDP is a mathematical framework for modeling sequential decision-making processes based on the Markov property \cite{howard1960dynamic}. An MDP can be defined with a set of states, $\mathcal{S}$, a set of actions, $\mathcal{A}$, a state transition function, $\mathcal{T}$, and the (immediate) reward, $\mathcal{R}(s,\alpha)$. We define policy as a mapping function from a given state to an action: $\pi : \mathcal{S} \rightarrow \mathcal{A}$. The objective of MDP is to find an optimal policy that maximizes the expected cumulative reward; i.e., $\pi_{\theta}^*(\alpha|s) = {\arg \max}_{\theta} \left[ \sum_{t=0}^{\infty} \gamma^t \cdot \mathcal{R}(s,\alpha) \right]$, where $\gamma \in [0,1]$ is the discount factor.

In reinforcement learning for autonomous driving and CACC, the design of the MDP, including state, action space, and reward function, is crucial for effective learning. A well-crafted state space provides essential information about the vehicle's environment and conditions, essential for realistic training. Our goal is to create a training environment that reflects real-world CACC complexities, aiding the RL agent in learning optimal driving policies for improved performance and safety. Upcoming sections will detail the state space, action space, and reward function we've developed for our CARL model.

\subsubsection{States and Observations}
In practice, it is challenging to define proper state space which can ensure \textit{cooperative} driving of multiple CACC vehicles. 
%
%
The state is a representation of the current environment in which the agent is living. 
It should be easy for an agent to observe the defined states, and the observed states should include all relevant information to take proper action. Likewise, in the longitudinal control system, we want the states to accurately reflect the current information about the vehicle while being as simple as possible so that it is easier to access during the following process \cite{kaelbling1996reinforcement, lin2020comparison}. However, it is sometimes difficult to achieve both conditions, which makes it even harder to define the state space properly.

One solution is to assume the \textit{partial observability} of the state space and approximate the policy with $\pi(\cdot | s) \approx \pi(\cdot | o)$, where the observation $o \in \mathcal{O}$ has a subset of the information of the actual state $s$. This approach is known as a Partially Observable MDP (POMDP), which is widely used in many real-world reinforcement learning problems in the transportation domain \cite{choi2021trajgail,yan2022unified}.

At time-step $t$, the observation of the $i$-th CACC vehicle, $o_i^t$, is defined as follows:
\begin{equation}\label{eq:obs}
  o_i^t = \left[ v_i^t, \alpha_i^t, d_i^t, \Delta v_i^t \right]^\top,
\end{equation}
where $v_i^t$ is the speed, $\alpha_i^t$ is the acceleration, $d_i^t$ is the spacing between the preceding vehicle and the ego vehicle, and $\Delta v_i^t = v_{i-1}^t - v_{i}^t$ is the relative speed between the preceding vehicle and the ego vehicle. 
{

Here we use local observations of each vehicle. Using local information in CACC brings practicality, as it doesn't require specialized sensors, ensuring seamless integration into existing ACC systems. It enhances robustness, reliability, and privacy, as it reduces susceptibility to communication issues and avoids sharing sensitive data. The approach is scalable, accommodating both CACC-enabled and non-CACC-enabled vehicles, while maintaining high performance and adaptability across various traffic conditions. Overall, leveraging local information makes our method an efficient and viable solution for CACC.
}
%




\subsubsection{Actions}
In previous reinforcement learning studies for CACC, longitudinal acceleration is used as the action. However, during implementation, we found that directly using the longitudinal acceleration often results in unstable training. Also, it is necessary to set arbitrary safety constraints to ensure safe driving without collision. This approach may limit the capability of learning optimal policy since the model is not able to learn proper actions at some specific region of state space (i.e., the CACC vehicle cannot learn the longitudinal control dynamics when they have to obey the safety constraint). 

{

Meanwhile, many car-following models based on machine learning have used prior knowledge for pre-training and achieved promising results. Some of them adopt physical-informed prior knowledge \cite{mo2021physics, ma2023physics}, which allows the model to better understand real-world physics and ensure safety. Others directly use pre-trained deep learning or deep reinforcement learning models \cite{zhu2018human, naing2022dynamic}, which has the advantage of greatly reducing training time.
}

%

Therefore, we give the vehicles some prior knowledge to get them trained faster. 
Specifically, we directly assist the action with a pre-defined car-following model and train the RL to adjust the acceleration to improve the output action of the vehicle with a pre-defined model. This has the advantage of ensuring a lower bound on model performance while reducing training uncertainty. In addition, using some models with safety restrictions (e.g. Intelligent Driving Model \cite{treiber2000congested} can also help the vehicle learn the safe driving strategy faster.


As a result, in this study, we use a base car-following model and learn how to adjust the base longitudinal control model. The base car-following model can be any car-following model or pre-trained deep learning/reinforcement learning model as long as it can output vehicle actions at each time step. The action $a$ is defined as the \textit{adjustment for longitudinal acceleration} of the ego vehicle ($\alpha_{i,adj} ^t$) from the longitudinal acceleration from the base car-following model ($a_{i,CF}^t$). The final acceleration ($a_i^t$) is calculated as:
%
%
\begin{equation}\label{eq:action}
  \begin{split}
  \alpha_{i}^t = \alpha_{i,CF}^t + \alpha_{i,adj}^t
  \end{split}.
\end{equation}




\subsubsection{Rewards}
{

Most of the reward function settings for RL-based ACC are similar because they aim to achieve common objectives and promote desired behaviors in the CACC system. These objectives typically include safety, comfort, efficiency, and traffic flow optimization. Common components of the reward function in RL-based CACC include:
\begin{itemize}
    \item {\textbf{Safety}: Encouraging the vehicle to maintain a safe distance from the leading vehicle to avoid collisions or unsafe following behaviors. Penalties are usually given for sudden braking or acceleration.}
    \item {\textbf{Comfort}: Promoting smooth and gradual acceleration and deceleration to ensure a comfortable ride experience for passengers.}
    \item {\textbf{Efficiency}: Rewarding the vehicle for maintaining steady speeds, minimizing unnecessary accelerations or decelerations, and achieving efficient fuel consumption.}
    \item {\textbf{Traffic Flow Optimization}: Encouraging the vehicle to follow the traffic flow and maintain a consistent speed to improve overall traffic stability and flow.}
\end{itemize}

These common components in the reward function align with the fundamental goals of ACC/CACC, which are to improve safety, comfort, and efficiency while maintaining smooth traffic flow. The reward function in most papers is a combination of the above aspects, and they are similar \cite{desjardins2011cooperative,zhu2018human, zhu2020safe, shi2021connected, jiang2022reinforcement}. While specific implementations of the reward function may vary depending on the system's requirements and objectives, the shared focus on achieving these common goals leads to similar reward function settings across many RL-based ACC studies. The convergence toward similar reward designs reflects the understanding of the essential characteristics and objectives of ACC control that researchers and practitioners aim to optimize through RL.
Therefore, in order to make a better comparison to verify our unique advantages after adding the Communication-Aware module, we used reward functions that had been validated several times in previous studies \mbox{\cite{zhu2020safe, zhu2018human}}. 
Following previous studies \cite{zhu2020safe}, we use a linear combination of three component reward functions as:
}
\begin{equation}\label{eq:reward}
  \begin{split}
  r_{i}^t = w_{g} r_{i,g}^t + w_{s} r_{i,s}^t + w_{c} r_{i,c}^t
  \end{split},
\end{equation}
where $r_{g}$, $r_{s}$, and $r_{c}$ are the component reward function representing gap, safety, and comfort, respectively, and $w_{g}$, $w_{s}$, and $w_{c}$ are the corresponding weights associated with each component reward function. $r_{i,*}$ represents each specific reward function and $\alpha_{*}$ represents the corresponding coefficient for the reward. For the specific reward design, the detailed definition of each reward is defined as follows:

\begin{equation}
     r_{i,g}^t = \frac{1}{{h_i^t}*\sigma \sqrt{2 \pi}} e^{\frac{-(\ln{h_i^t}-\mu)^2}{2 \sigma^2}} ,
\end{equation}

\begin{equation}
  r_{i,s}^t  = \begin{cases}
    \log \left(\frac{{TTC}_i ^t}{4} \right)  & , \quad \text{if 0 $<$ TTC $\leq$ 4 } ,\\
    0  & , \quad \text{otherwise}\\
  \end{cases}
\end{equation}

\begin{equation}
  \begin{split}
  r_{i,c}^t  = - \left(\frac{{j_i^t}}{ a_{i,max} - a_{i,min}}\right)^2 ,
  \end{split}
\end{equation}
where $v_i$ represents the speed of the $i$-th vehicle, $a_i^t$ represents the acceleration of the $i$-th vehicle, and $j_i$ represents the jerk of the $i$-th vehicle. $a_{i,max}$ is the maximum acceleration rate ($a_{i,max}>0$), and $a_{i,min}$ is the maximum deceleration rate ($a_{i,min}$<$0$). $TTC$, \textit{Time-to-Collision}, is one of the widely-used safety surrogate measures. $h_i$ is the headway of the $i$-th vehicle, which represents the duration between vehicles measured in time. $\sigma$ and $\mu$ are two fixed coefficients. $h_i$, $TTC$ and $j_i$  are defined as follows:

\begin{equation}
h_i^t = \frac{x_{i-1}^t - l_{i-1} - x_{i}^t }{v_{i}} ,
\end{equation}

\begin{equation}
  \begin{split}
  TTC_i^t = \frac{x_{i-1}^t - l_{i-1} - x_{i}^t }{v_{i}^t - v_{i-1}^t}, & \quad \text{if $v_{i}>v_{i-1}$} ,\\
  \end{split}
\end{equation}
\begin{equation}
    j_i^t = \frac{a_i^t - a_i^{t-1}} {\Delta t} ,
\end{equation} 
where $x_{i-1}^t$ is the position of the preceding vehicle at time $t$, $x_{i}^t$ is the position of the ego vehicle at time $t$, and $l_{i-1}$ is the length of the preceding vehicle. $a_{i}^t$ is the acceleraltion of the $i$-th vehicle at time $t$. $\Delta t$ is the interval between two actions.

\subsection{Communication-Aware RL with Message Processing}

Considering the characteristics of the CACC communication system, we designed a unique communication structure for CACC called the Communication Aware (CA) Module. The proposed model is based on the V2V communication, where each vehicle in our system communicates directly with the surrounding vehicles.

An important point of vehicle communication is how to process the received information. Instead of simply using direct physical information (such as velocity, position, acceleration, etc.), we use a set of neural networks to extract high-dimensional features of the received information. The networks in the CA model are used for their ability to process and extract features from vehicle communication data. This network efficiently handles inputs from surrounding vehicles in the CACC system, transforming this data through its multiple interconnected layers. Its implementation within the CA module significantly enhances the CACC system's ability to interpret and utilize vehicle-to-vehicle communication for improved decision-making and overall system performance.

\begin{figure*}[t]
 \centering
 \includegraphics[width=0.9\textwidth]{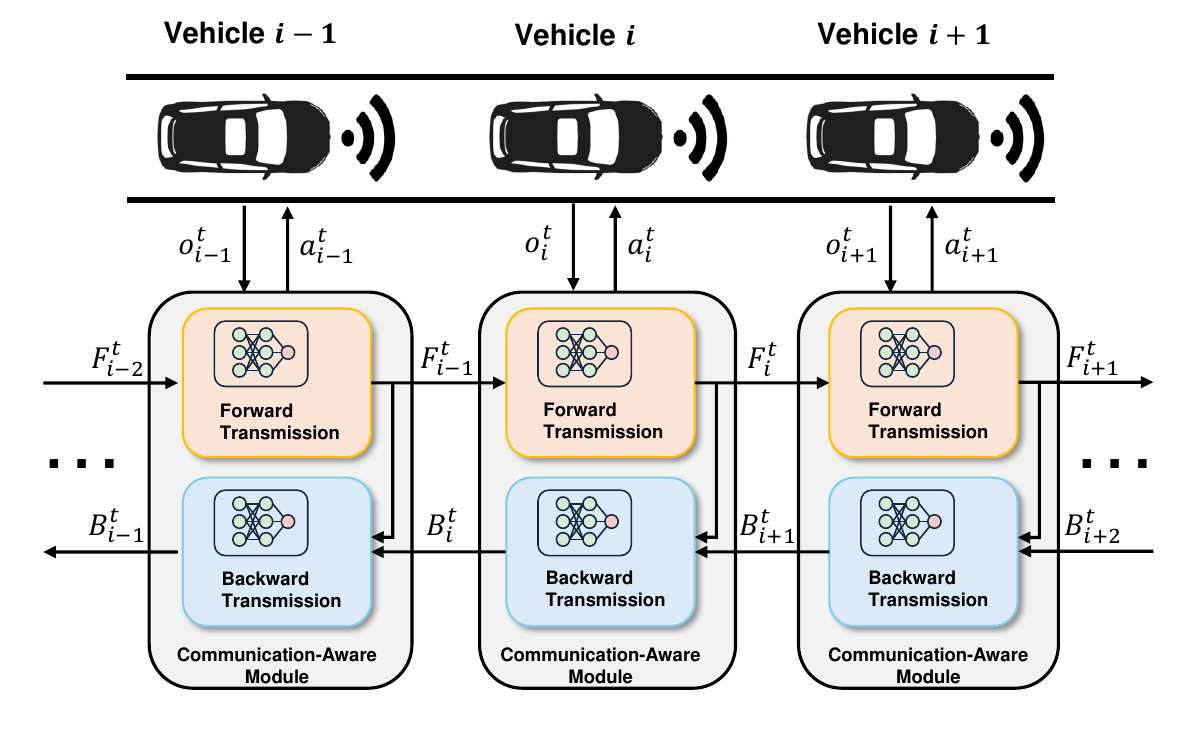}
 \caption{Architecture of Communication-Aware Module: The communication module receives the information from the front and rear cars and then processes the information using a network module, after which the information is then output to the surrounding vehicles. The output of the current action is given using the obtained information together with the RL network.}
 \label{Architecture}
\end{figure*}


In the proposed module, one network is responsible for transmitting information from the following car to the proceeding car (forward transmission), while the other is responsible for transmitting information from the proceeding car to the following car (backward transmission). Figure \ref{Architecture} shows the structure of our model. $F^t_i$ is the forward transmission message from the $i$-th vehicle to the preceding vehicle at time $t$. $B^t_i$ is the backward transmission message from the $i$-th vehicle to the following vehicle at time $t$. At each time $t$, the $i$-th vehicle collects the observation, $o_i^t$ from the environment. By combining the forward transmission message from the following vehicle ($(i+1)$-th vehicle), $F_{i+1}^t$, and the observation, the ego vehicle generates the forward transmission message as: 
\begin{equation}
  \begin{aligned}
  F_{i}^t = f_{F} \left(o_i^t, F_{i+1}^t \right)
  \end{aligned},
\end{equation}
where $f_F$ is the forward transmission network.
Then, the forward transmission message is combined with the backward transmission message from the leading vehicle ($(i-1)$-th vehicle), $B^t_{i-1}$, to generate the backward transmission message from the ego vehicle as follows:
\begin{equation}
  \begin{aligned}
  B_{i}^t = f_{B} \left(F_{i}^t, B_{i-1}^t \right)
  \end{aligned},
\end{equation}
where $f_{B}$ is the backward transmission network.

{


The forward transmission represents the process by which information is communicated from the following vehicle to the current vehicle. This information exchange allows the current vehicle to gain insights into the behavior and intentions of its leading vehicle, enabling it to make informed decisions about speed and acceleration adjustments.

Conversely, backward transmission involves the communication of information from the preceding vehicle to the current vehicle. This backward information flow enables the following vehicle to receive updates on the current vehicle's actions and intentions, facilitating coordinated actions within the CACC system and maintaining safe following distances.
In CACC car following, the vehicle often needs to make different control responses to the preceding and following vehicles. Treating forward and backward transmission as distinct processes allows the algorithm to capture this asymmetry and enables the ego-vehicle to learn more detailed and differentiated control strategies based on communication information. This differentiation in control strategies, in turn, enhances the efficiency and effectiveness of the CACC system.

In our architecture, the order of forward transmission followed by backward transmission is utilized. However, it is important to recognize that when considering the nature of two-way communication, the order in which forward and backward transmissions occur does not have an impact on the actual performance of CACC.
}



\subsection{Implementation with Actor-Critic Network}
\begin{figure*}[t]
 \centering
 \includegraphics[width=1.0\textwidth]{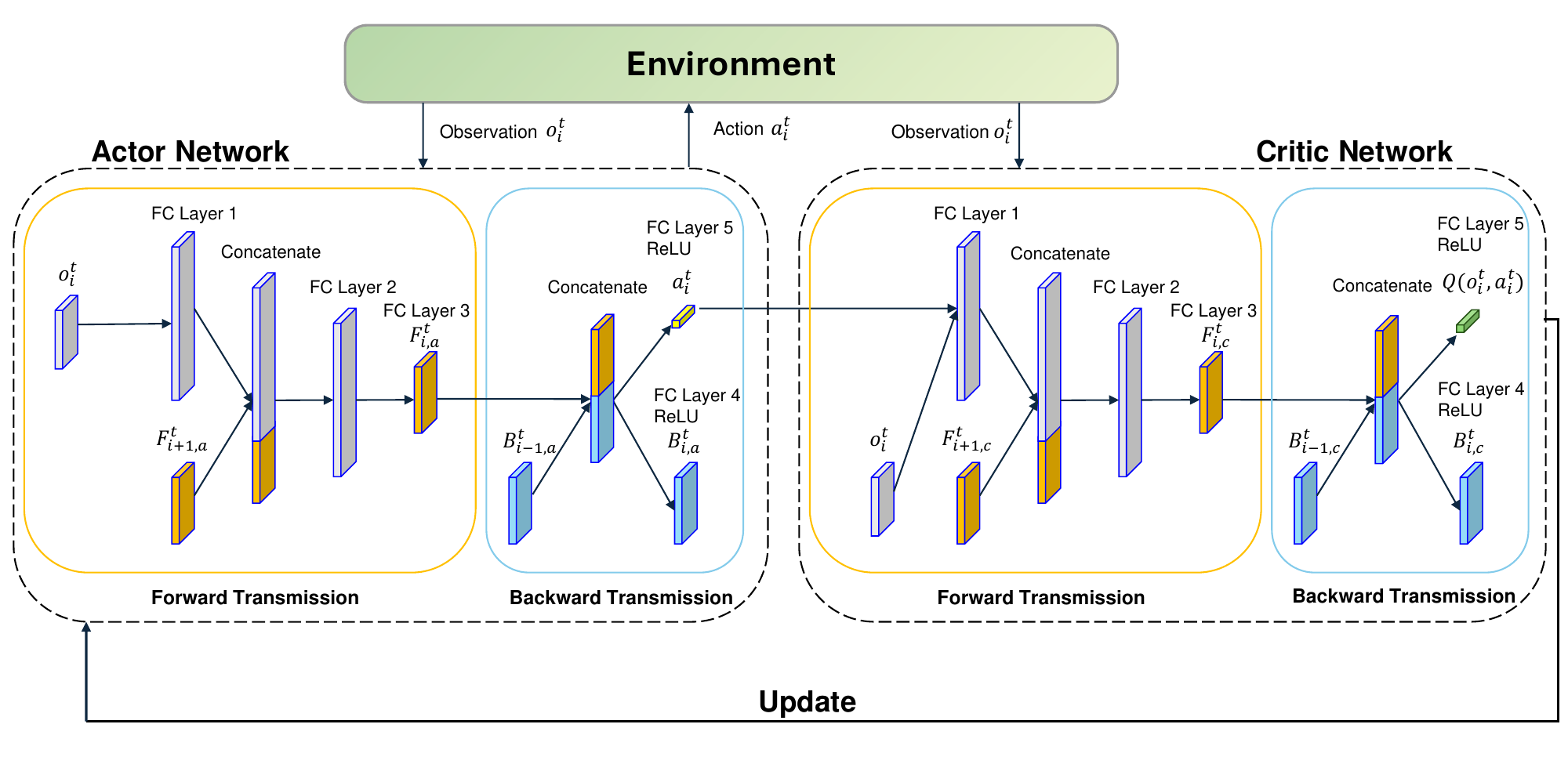}
 \caption{Communication-Aware RL framework combined with Actor-Critic network}
 \label{Network}
\end{figure*}

Figure \ref{Network} shows how we combined the Communication-Aware module with the actor-critic network. Actor-critic is a classic reinforcement learning algorithm that combines policy-based and value-based approaches to improve the learning efficiency and stability of the agent. The actor represents the policy or the agent's decision-making function that chooses actions based on the current state of the environment. The critic, on the other hand, estimates the value of the policy by providing feedback to the actor on how good its actions were in a particular state. The critic uses the temporal difference (TD) learning algorithm to learn the value function, which is a measure of how good a state or action is in terms of achieving the agent's goal. The actor then uses this value function to update its policy by adjusting the probability distribution over actions to maximize the expected cumulative reward. By combining the policy-based and value-based approaches, the actor-critic can learn more efficiently and reliably than either approach alone.

In this actor-critic model, we have two sets of message systems. The message processed and passed by the actor network will only be received by the actor-network of the front and rear vehicles. Similarly, the message processed and passed by the critic network will only be received by the critic network of the front and rear vehicles.

In the actor network, the forward transmission network receives the environment observation information $o_i^t$ and the actor message $F_{i+1,a}^t$ from the rear vehicle. Through several fully connected layers, the forward transmission network outputs the forward actor message $F_{i,a}^t$, which will be passed to the front vehicle and the backward transmission network. The backward transmission network receives the backward actor message $B_{i-1,a}^t$ from the front vehicle and $F_{i,a}^t$ sent by the forward transmission network. After fusing the forward message $F_{i,a}^t$ and backward message $B_{i-1,a}^t$, the backward transmission network generates the action $a_i^t$ and the backward actor message $B_{i,a}^t$. The action $\alpha_t^t$ will be sent to the environment to control the vehicle and also to the critic network for updating the network, while the message $F_{i,c}^t$ will be passed to the actor-network of the rear vehicle.

In the critic network, most of its structure is similar to the actor network. The inputs of the forward transmission network are the observation $o_i^t$, the action $a_t^t$, and rear vehicle's forward critic message  $F_{i+1,c}^{t}$.  Then the forward transmission network outputs forward critic message $F_{i,c}^t$ to the backward transmission network. The backward transmission network receives message $F_{i,c}^t$ and $B_{i-1,c}^t$, while generates the backward critic message $B_{i,c}^t$ and state-value function $Q(o_i^t,a_i^t)$.

{

Considering the different roles of actors and critics in the network, actor messages $F_a, B_a$, and critic messages $F_c, B_c$ do not need to be consistent. The actor message uses the critic's guidance to improve its actions, while the critic message uses the actor's actions to evaluate the policy's performance. This combination accelerates learning and leads to a more robust and reliable policy.
}

In each iteration of training, the action is generated by the actor-network, while the critic-network outputs the state-value function to update the parameters of the action network by gradient descent. The updating process is as follows. To update the critic network, we first calculated the value function $y_i^t$:


\begin{equation}  
y_i^t=r_i^t+\gamma Q\left(o_{i+1}^{t+1}, \pi\left(o_{i+1}^{t+1} \mid \theta^\pi \right) \mid \theta\right),
\end{equation}
where $r_i^t$ is the reward received after taking the action $a_i^t$ in state $o_i^t$. $\gamma$ is the discount factor, which determines the importance of future rewards. $o_{i+1}^{t+1}$ is the next observation of the state after taking the action a. $Q$ is the Q-value function to estimate the reward of taking action. $\pi$ is the policy of network, $\theta$ is the parameters of the policy.

Then we update the critic network by minimizing loss function $L$:
\begin{equation}
L=\frac{1}{N} \sum_{i=1}^N\left(y_i^t-Q\left(s_i^t, a_i^t \mid \theta\right)\right)^2,
\end{equation}
where $s_i^t$ is the states of agent $i$ at time $t$, in a partially observed environment, states are equivalent to observations.
And the actor network is updated by calculating the gradient of Q-value functions:
\begin{equation}
\left.\nabla_{\theta } J \approx \frac{1}{N} \sum_{i=1}^N \nabla_a Q \left(s_i^t, a_i^t \right)\right|_{s_i^t=o_i^t, a_i^t=\pi_\theta \left(o_i^t\right)},
\end{equation}
where $\pi_\theta$ represents the parameters $\theta$  under policy $\pi$.



It is important to note that our CA module is not only able to combine with policy-based RL algorithms such as actor-critic, but it also can combine with a variety of other RL algorithms. It only requires modifications to the input and output of the network in the RL framework, which makes it very flexible in applications. In the experiments of the next section, we combine it with DDPG and TD3, which are two typical RL algorithms.

\section{Experimental Setting}\label{sec:expset}

\subsection{Dataset}
The NGSIM dataset, collected for the "Next Generation Simulation" project, includes diverse traffic scenarios from four U.S. regions. It's been post-processed to provide vehicle trajectory data, essential for vehicle-road collaboration research. Given that most ACC usage occurs on highways, we've chosen highway trajectory data from NGSIM, particularly from California's I-80 freeway, for testing different ACC algorithms.

For training and testing our reinforcement learning model, referring to the methods of previous researchers \cite{zhu2020safe}, we selected 1400 vehicle-following trajectories, using 70\% for training and 30\% for testing. These data, from April 13, 2005, offer high-accuracy vehicle location information, making them ideal for our study.


\subsection{Simulation Environment}
In each simulation training iteration, we randomly select a vehicle trajectory from the processed NGSIM dataset as the first leading vehicle. Then there are $N$ CACC vehicles following the preceding vehicle of its own. At the time-step $t$, the acceleration of the CACC vehicle is determined by the baseline longitudinal control model and actions from the RL model as discussed in Equation \ref{eq:action}. The speed and the position of each vehicle are updated by the following equation:

\begin{equation}\label{eq:update}
  \begin{split}
  & v_i^{t+\Delta t} = v_i^t + \alpha_i^t \Delta t = v_i^t + \left(\alpha_{i,CF}^t + \alpha_{i,adj}^t \right) \Delta t \\
  & x_i^{t+\Delta t} = x_i^t + v_i^{t+\Delta t} \Delta t,\\
  \end{split}
\end{equation}

\noindent
where $\Delta t$ is the unit of time-step, which is defined as 0.1 $s$.

The baseline longitudinal control model used for Equation \ref{eq:action} is the Intelligent Driver Model (IDM) \cite{kesting2010enhanced}. IDM is one of the most widely-used longitudinal control models for microscopic traffic simulation. The equation for calculating the acceleration of the ego vehicle in IDM is defined as:

\begin{equation}\label{eq:IDM}
  \begin{split}
    & \alpha^t = \alpha\left( 
    1 - 
    \left(\frac{v^t}{v_0}\right)^\delta - 
    \left( \frac{s^*(v^t, \Delta v^t) }{s^t} \right)^2
    \right) \\
    & s^*(v^t, \Delta v^t) = s_0 + v^t T + 
    \frac{v^t \Delta v^t}{2 \sqrt{\alpha \beta}}\\
  \end{split},
\end{equation}

\noindent
where $\alpha$ is the maximum vehicle acceleration, $\beta$ is the comfortable braking deceleration, $v_0$ is desired speed, and $s^*$ is the desired gap. In the original definition of IDM, $s^t$ is the spacing between the front vehicle, and $\Delta v^t$ is the relative speed at the timestep $t$. The parameter value settings of IDM are as follows:  $\alpha$ = 3 $m/s^2$ , 
    $\beta$ = 2 $m/s^2$ ,
    $v_0$ = 120 $km/hr$ ,
    $s_0$ = 2 $m$ ,
    $T$ = 1.5 $s$ ,
    $\delta$ = 4 .
%
%
We use the parameters based on the original calibration of IDM for the NGSIM dataset \cite{chen2010calibration}.




\subsection{Baseline Models}

\begin{itemize}
  \item \textbf{IDM} --- 
  The intelligent driver model (IDM) is a mathematical model that describes and predicts the behavior of vehicles in traffic. It is based on a set of rules that determine how a driver should accelerate or decelerate based on the surrounding traffic conditions.
  \item \textbf{Krauss} --- Krauss model is a microscopic, space-continuous, longitudinal control model based on vehicles' speed. It defined a safe speed as follows:
  \begin{equation}
v_{\text {safe }}=v_l(t)+\frac{g(t)-v_l(t) t_r}{\frac{v_l(t)+v_f(t)}{2 \beta}+t_r}
\end{equation}
where $v_l(t)$ is the speed of the leading vehicle in time $t, g(t)$ is the gap to the leading vehicle in time t, $t_r$ is the driver's reaction time and $\beta$ is the maximum deceleration of the vehicle.
  \item \textbf{DDPG} --- 
  As mentioned in Section 2, Deep Deterministic Policy Gradients (DDPG) is a model-free, off-policy reinforcement learning algorithm for learning continuous control policies. 
  \item \textbf{TD3} --- 
  Twin delayed deep deterministic policy gradient (TD3) is an improved version of DDPG \cite{fujimoto2018addressing}. 
  \item \textbf{MADDPG} ---
Multi-Agent Deep Deterministic Policy Gradient (MA-DDPG) is a variant of the DDPG algorithm that is designed for MARL settings \cite{lowe2017multi}, where multiple agents interact with each other in a shared environment. In MADDPG, each agent maintains its own policy and Q-value function and learns from its own experiences as well as from the experiences of the other agents. 
  \item \textbf{CA-DDPG} --- 
  As mentioned in Section 3, we combined our CARL module with DDPG and developed a novel reinforcement algorithm, which is called Communication-Aware DDPG (CA-DDPG).
  \item \textbf{CA-TD3} --- 
   Similar to CA-DDPG, CA-TD3 is the combined version of the Communication-Aware module and TD3.
\end{itemize}

\subsection{Evaluation Metrics}
As aggregated measures, we used five metrics to measure the performance of each model: Headway, Jerk, Speed, Counts of TTC $<$ 4, Counts of TTC $<$ 1.5 and Dampening Ratio.

Headway represents the distance or duration between vehicles in a transit system measured in time. The transit system is more efficient when its value is between 1s and 2.5s \cite{zhang2007examining}. Speed is the average speed of the overall traffic flow, under the condition of the speed limit, we want the traffic to keep the speed as high as possible. Jerk is a comfort index defined according to the previous section, and the smaller its value, the more comfortable it is. TTC stands for Time to collision. Research shows that 1.5 seconds and 4 seconds are two critical values of TTC that affect traffic safety \cite{minderhoud2001extended}. 

Dampening ratio is a measure of the string stability of the CACC control algorithm. String stability refers to the ability of a platoon of vehicles to maintain a stable configuration as they follow each other in a convoy. In other words, it means that the distances between the vehicles in the platoon remain constant over time, and the platoon as a whole behaves in a coordinated manner. It is an important property for ACC systems, as it ensures that the system can operate safely and efficiently in real-world traffic conditions. A lack of string stability can lead to traffic congestion, safety risks, and reduced system performance. The dampening ratio $d_{p}$ is calculated as follows:

\begin{equation}
d_{p}=\frac{\left\|a_i^t\right\|_2}{\left\|a_0^t\right\|_2}=\frac{\left(\sum_{t=0}^N\left|a_i^t\right|^2\right)^{\frac{1}{2}}}{\left(\sum_{t=0}^N\left|a_0^t\right|^2\right)^{\frac{1}{2}}},
\end{equation}
where $N$ denotes the time length and $a_i^t$ means the acceleration of vehicle $i$ at time $t$. $i$ is the index of the following vehicle, and index 0 represents the leader vehicle.

\section{Results}\label{sec:result}

This section shows the testing results of different models (IDM, Krauss, MADDPG, DDPG, CA-DDPG, TD3, CA-TD3) under the NGSIM dataset. Table \ref{metrics1} shows these models' average values of Headway, Jerk, Speed, Dampening Ratio, Counts of TTC $<$ 4, and Counts of TTC $<$ 1.5.

\subsection{Aggregated Measures}

\begin{figure*}[t]
     \centering
     \includegraphics[width=1.0\textwidth]{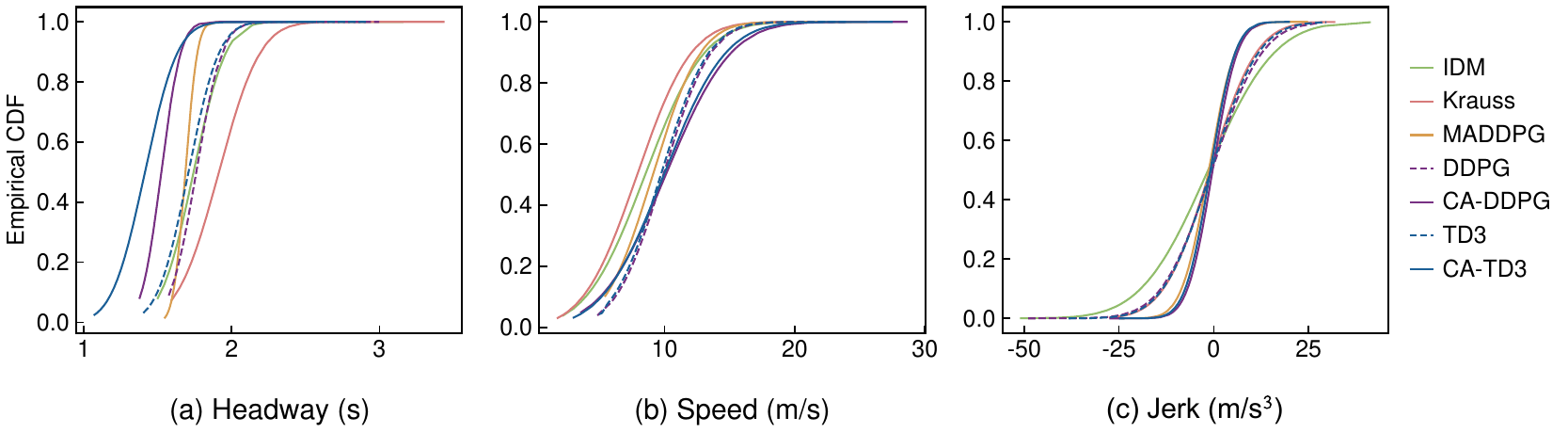}
     \caption{Empirical cumulative distribution of different models in speed, headway and jerk}
     \label{edf}
\end{figure*}

\subsubsection{Headway}
Our CA-DDPG model excels in headway, with a median around 1.4s, as shown in Figure \ref{edf} (a). It rarely exceeds 2.5s, indicating minimal inefficient control, thanks to our Communication-Aware module. In contrast, models like IDM, DDPG, Krauss, and MADDPG have a broader headway range, impacting following efficiency.

\subsubsection{Jerk}
CA-DDPG boasts superior comfort with a jerk value of 0.381 m/s$^3$, lower than traditional models and a significant improvement over standard DDPG. MADDPG is slightly better in comfort but requires more complex communication than CA-DDPG.

\subsubsection{Speed}
Figure \ref{edf} (b) shows CA-DDPG achieving the highest average speed at 10.256 m/s. The Communication-Aware module enhances DDPG's speed from 9.819 m/s, outperforming IDM and Krauss, and boosts overall traffic flow efficiency.

\subsubsection{Counts of TTC under the threshold}
Balancing efficiency and safety is crucial in CACC. Faster speeds often lower Time-To-Collision (TTC), increasing accident risks. Our CA-DDPG and CA-TD3 models effectively combine speed and safety, improving upon the safety of DDPG and TD3 while slightly increasing speed.

\subsubsection{String Stability}
Table \ref{metrics1} indicates that IDM and Krauss have higher dampening ratios, showing less capability to handle speed oscillations. DDPG and TD3 perform better in string stability. The inclusion of the CA module in CA-DDPG and CA-TD3 further enhances this, significantly improving string stability in CACC traffic flow.

\begin{table*}[b]
    \centering
    \begin{tabular}{c|c|c|c|c|c|c}
    \toprule
        \textbf{Model} & \textbf{Headway (s)}  & \textbf{Jerk (m/s$^3$}) & \textbf{Speed (m/s)} &\textbf{TTC $<$ 4 (s)} & \textbf{TTC $<$ 1.5 (s)} & \textbf{Dampening Ratio} \\  \midrule
        \textbf{IDM}  & 1.795  & 0.515 & 8.756  & 201.6 & 78.7 & 0.667\\ 
        \textbf{Krauss} & 1.988 & 0.453  & 7.958 &  239.6 & 66.2  & 0.623\\ 
        \textbf{MADDPG}  & 1.546 & 0.369 & 9.365  & 298.6 & 86.6 & 0.492\\ 
        \midrule
        \textbf{DDPG}    & 1.720 & 0.481  & 9.819  & 280.6 & 117.2 & 0.546\\ 
        \textbf{CA-DDPG} &\bfseries 1.517  &\bfseries 0.381  &\bfseries 10.256   &\bfseries  106.3 & \bfseries61.2 & \bfseries0.498 \\ 
        \midrule
        \textbf{TD3}    & 1.707 & 0.452  & 9.716  & 213.5 & 99.8 & 0.568\\ 
        \textbf{CA-TD3} & \bfseries 1.412 &  \bfseries 0.316 &\bfseries 9.939  & \bfseries113.7 & \bfseries 55.1 & \bfseries 0.465\\ 
        \bottomrule      
    \end{tabular}
     \caption{Average Headway, Speed, Jerk, TTC values, and Dampening Ratio for each model. }
    \label{metrics1}
\end{table*}



\begin{figure*}[]
 \centering
 \includegraphics[width=1.0\textwidth]{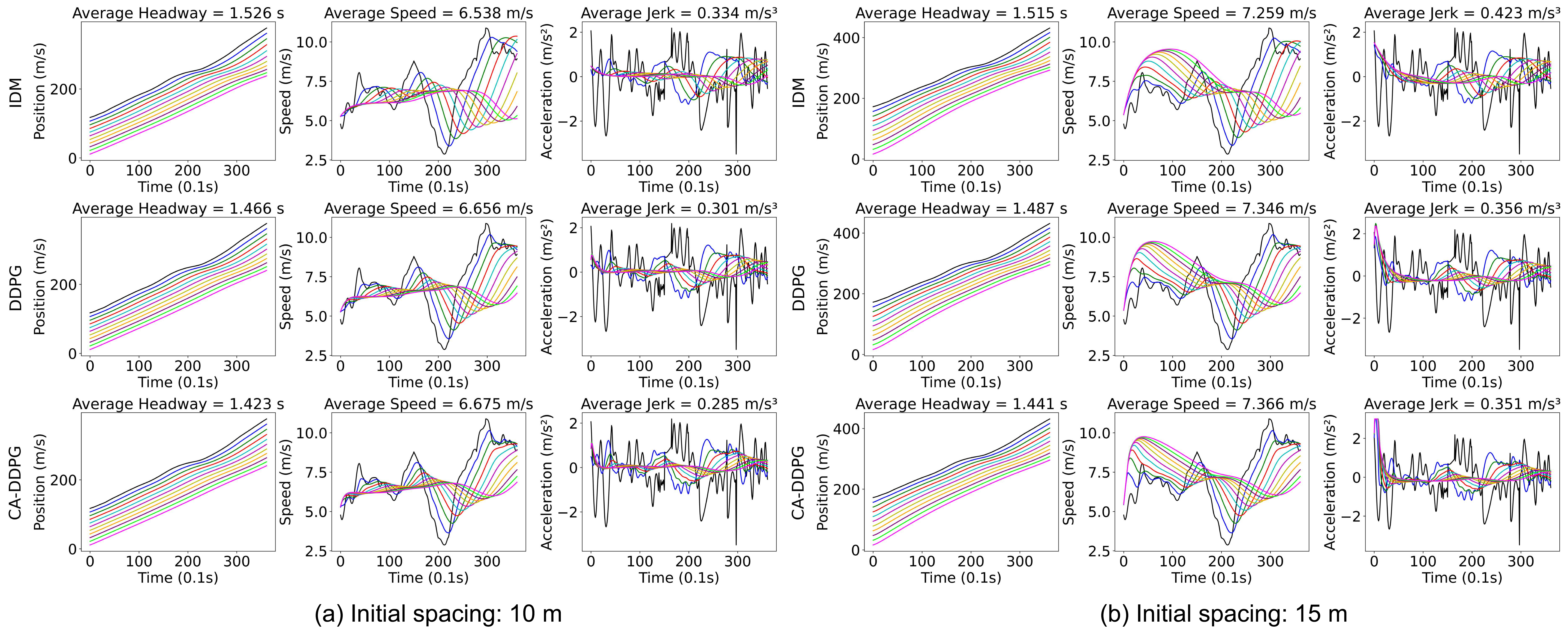}
 \caption{Position, speed, acceleration versus time step t in an NGSIM trajectory for different models with different initial spacing settings}
 \label{stg10}
\end{figure*}

\subsection{Trajectory based Performance}

Figure \ref{stg10} shows how different models follow a leader under varying initial spacing. In scenarios where the leader sharply decelerates (Figure \ref{stg10} (a)), the IDM model's conservative strategy results in lower speeds and delayed reactions, potentially causing inefficiency and congestion. DDPG and CA-DDPG offer improved comfort, with CA-DDPG responding quickly to the leader's actions and maintaining higher, stable speeds, demonstrating superior string stability.

With larger initial spacing (Figure \ref{stg10} (b)), vehicles generally move faster. IDM still performs the slowest, especially during sharp decelerations by the leader. DDPG shows quicker reactions than IDM but struggles at higher speeds. CA-DDPG stands out with optimal speed, comfort, and safety, efficiently adjusting to the leader’s speed changes.







\subsection{Generalization}
Due to the environment-sensitive nature of deep reinforcement learning, pre-trained deep reinforcement learning algorithms tend to have poor generalization ability \cite{kirk2021survey}. For multi-agent reinforcement learning, it is more sensitive to the state and the number of agents. Therefore, we tested the CARL model under a different number of vehicles / unknown scenarios and compared it with other algorithms to test its generalization capability.

\begin{table*}[!ht]
    \centering
    \resizebox{\textwidth}{!}{
    \begin{tabular}{c|c c||c|c|c|c|c|c}
    \toprule
        \textbf{Model} &\textbf{Train} &\textbf{Test} & \textbf{Headway (s)}  & \textbf{Jerk (m/s$^3$}) & \textbf{Speed (m/s)} &\textbf{TTC $<$ 4 (s)} & \textbf{TTC $<$ 1.5 (s)} &\textbf{Dampening Ratio} \\  \midrule
        \textbf{IDM}  & - & 10 & 1.795 & 0.515 & 8.756 & 201.6 & 78.7 & 0.667 \\ 
        \textbf{IDM}  & - & 20 & 1.721 & 0.653 & 8.249 & 355.9 & 194.2 & 0.713\\
        \textbf{DDPG}  & 1 & 10 & 1.720 & 0.481 & 9.365 & 239.6 & 66.2 &0.546\\ 
        \textbf{DDPG}  & 1 & 20 & 1.832 & 0.550 & 8.941 & 273.9 & 84.7 &0.581\\ 
        \textbf{MADDPG}  & 10 & 10  & 1.546 & 0.369 & 9.365 & 298.6  & 86.6 &0.492\\ 
        \textbf{MADDPG}  & 20 & 20 & 1.532 & 0.398 & 9.136 & 325.1 & 117.3 &0.569\\ 
        \midrule
        \textbf{CA-DDPG}  & 10 & 10 & 1.517 & 0.381 & 10.256 & 106.3 & 61.2 & 0.498\\ 
        \textbf{CA-DDPG}  & \bfseries 10 & \bfseries 20 & \bfseries 1.598 & \bfseries 0.403 & \bfseries 9.985 & \bfseries 189.2 & \bfseries 113.1 &\bfseries 0.521\\ 
        \bottomrule
    \end{tabular}
    }
     \caption{Average Headway, Speed, Jerk, TTC values for each model under 20 vehicles. The values in parentheses are their performance under 10 vehicles.}
      \label{20Vehicles}
\end{table*}

\subsubsection{Generalization for the number of vehicles}
Traditional single-agent reinforcement learning algorithms like DDPG and TD3 struggle in multi-agent systems such as CACC due to their inability to facilitate communication and coordination among agents. In contrast, multi-agent reinforcement learning algorithms, like MADDPG, are designed for better performance in such systems by enabling coordination among agents. However, they face challenges with scalability, as their complexity grows with the number of agents, making them less suitable for large-scale systems.

Our proposed CARL algorithm addresses these limitations by combining the strengths of both single-agent and multi-agent approaches. It adapts single-agent algorithms for multi-agent contexts, allowing each agent to communicate and coordinate with others during training. This is achieved through a communication module integrated into the CARL, facilitating shared policy updates among agents.

To evaluate the generalization ability of our model, we conducted experiments with varying numbers of vehicles in the CACC system. We trained the single-agent DDPG algorithm with one vehicle and tested it with larger groups. For MADDPG, we trained and tested with the same number of agents due to its limited scalability. In contrast, our CA-DDPG, with its shared policy, was trained with 10 agents and successfully generalized to both 10 and 20 agents, demonstrating its superior adaptability in multi-agent systems.

\begin{figure*}[!htb]
 \centering
 \includegraphics[width=0.9\textwidth]{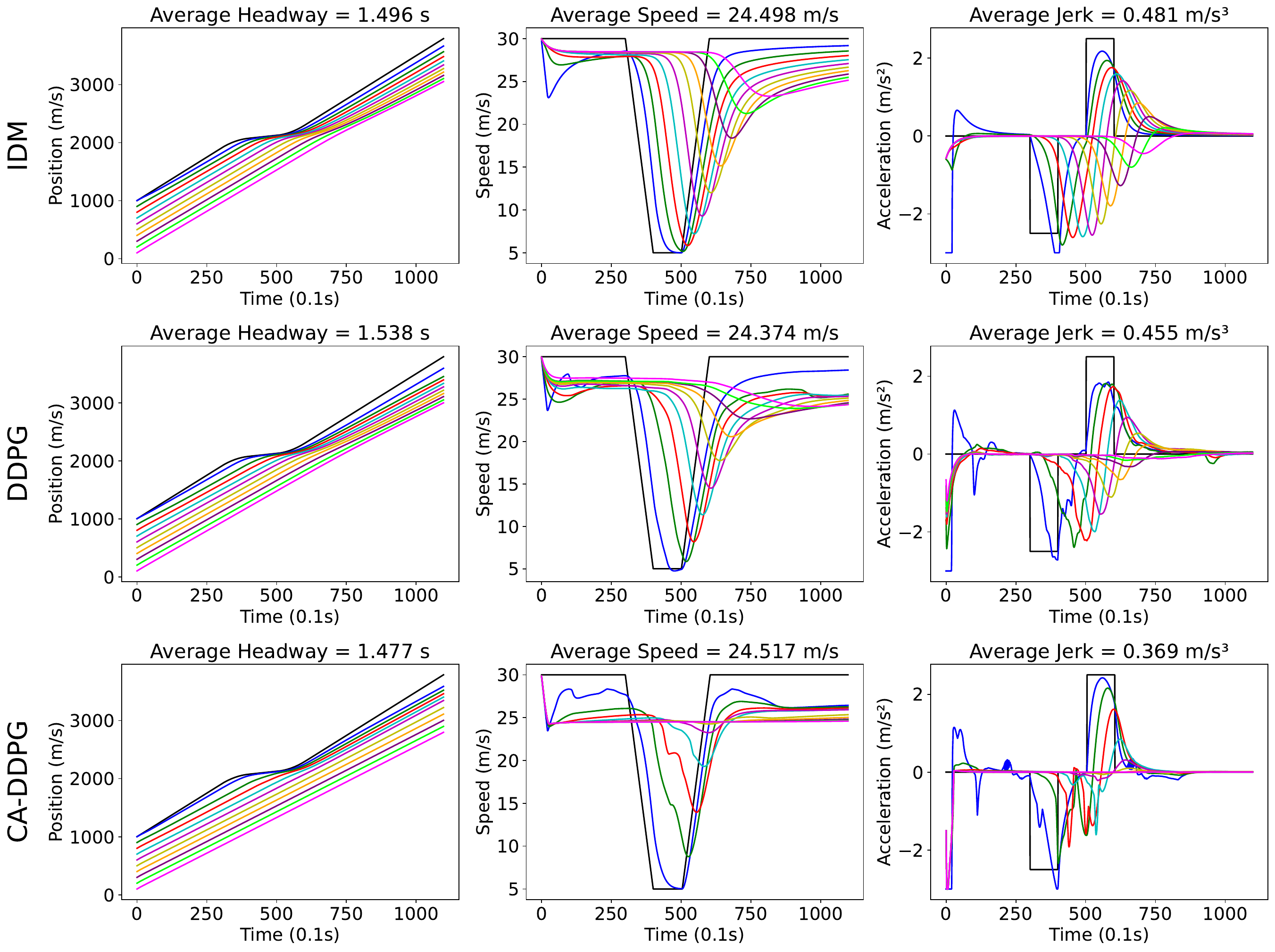}
 \caption{Position, speed, acceleration versus time step t for different models under stop-and-go scenario.}
 \label{TrajSG}
\end{figure*}



In the scenario with higher vehicle density, as shown in Table \ref{20Vehicles}, traditional algorithms like IDM and DDPG significantly declined in performance metrics such as headway, speed, and safety, with IDM also showing an increase in risky driving behaviors. MADDPG, trained and tested with 20 vehicles, performed better, but all three algorithms (IDM, DDPG, and MADDPG) struggled with increased Dampening Ratio, indicating difficulty in adapting to speed oscillations of the front vehicle.In contrast, our CA-DDPG algorithm demonstrated more resilience in this scenario, showing less degradation in key metrics and maintaining better string stability. This suggests that CA-DDPG's integration of communication and coordination in a multi-agent framework equips it more effectively to handle complex traffic situations with high vehicle density.





\subsubsection{Generalization experiments for unknown scenarios}
RL often struggles in unfamiliar environments, hindered by its need for environmental interactions to learn optimal decisions. Without prior knowledge, RL agents face a difficult exploration-exploitation dilemma, balancing the gathering of new information against utilizing existing knowledge for maximum rewards. This balance is crucial and challenging, directly affecting performance in new settings.

RL agents also grapple with poor generalization in unfamiliar scenarios, leading to suboptimal responses to new environmental conditions. This issue is particularly relevant in CACC, where vehicles regularly encounter varied traffic situations. However, our CARL algorithm, enhanced by V2V communication, is designed to perform effectively even in these unknown CACC scenarios.

To assess this, we conducted experiments simulating a Stop-and-go longitudinal control scenario with significant speed changes - a situation not covered in the NGSIM dataset and thus novel to both DDPG and CA-DDPG. This test aims to evaluate the algorithms' adaptability and performance in unfamiliar traffic conditions.

\begin{table*}[!tb]
    \centering
    \begin{tabular}{c|c|c|c|c|c|c}
    \toprule
        \textbf{Model} & \textbf{Headway (s)}  & \textbf{Jerk (m/s$^3$}) & \textbf{Speed (m/s)} &\textbf{TTC $<$ 4} & \textbf{TTC $<$ 1.5} & \textbf{Dampening Ratio} \\  \midrule
        \textbf{IDM}  & 1.496  & 0.481 & 24.498  & 6 & 2 & 0.683\\ 
        \textbf{DDPG}    & 1.538 & 0.455 & 24.374  & 8 & 1& 0.451 \\ 
        \textbf{CA-DDPG} & \bfseries1.477  & \bfseries0.369 & \bfseries24.517   & \bfseries 4 & \bfseries0 & \bfseries0.332\\ 
        \bottomrule
    \end{tabular}
     \caption{Average Headway, Speed, Jerk, counts of TTC and Dampening Ratio values for each model in extreme stop-and-go scenario. }
     \label{SaG}
\end{table*}


In the stop-and-go scenario, Table \ref{SaG} and Figure \ref{TrajSG} show that while IDM maintains stable performance, DDPG, a single-agent RL algorithm, underperforms, indicating poor generalization. DDPG surpasses IDM in string stability but lags in speed and comfort. Both IDM and DDPG struggle with velocity oscillations from the lead vehicle.

Contrastingly, CA-DDPG, enhanced with a communication module, responds more effectively to sudden braking by the leader, quickly mitigating speed oscillation. This swift response leads to CA-DDPG outshining both IDM and DDPG in speed, headway, comfort, safety, and string stability. These results highlight the robustness and strong generalization ability of our CARL model in challenging traffic conditions.

\section{Conclusion and Discussion}\label{sec:conclusion}

{

In this paper, we introduce a novel Communication-Aware module combined with Reinforcement Learning (CARL) to improve longitudinal control in CACC. Our method integrates a V2V communication mechanism, enabling vehicles to leverage information from others for enhanced decision-making. A key feature is the networks within the Communication-Aware module, which processes high-dimensional data for effective information extraction. This allows vehicles to obtain general traffic flow information while maintaining a consistent policy.

We validated CARL against standard CACC algorithms using real-world NGSIM datasets, where it demonstrated superiority in speed, comfort, safety, and robust string stability. Our approach also shows strong generalization in various scenarios and demonstrates advantages in scalability and stability.

For future works, we will focus on the following three aspects:

Enhancing CARL's Adaptability: Future developments in CARL will focus on increasing its adaptability to various road conditions and traffic patterns. This includes tailoring the system to respond dynamically to different environmental factors like weather, road types, and traffic densities.For example, under different weather conditions, the CA module can perform a series of fine-tuning to process data with different emphasis to ensure safety and efficiency.

Extending CARL's Application Range: Another key area of focus will be extending CARL's applications to cover a broader spectrum of tasks for CAVs. The characteristics of CARL allow it to be integrated into RL algorithms for other tasks such as lane changing, and collision avoidance. By expanding CARL's capabilities, we aim to create a more intelligent control system that can handle various driving tasks, contributing to the overall autonomy of CAVs.

Exploring Robustness: Future research will focus on enhancing CARL's scalability and robustness in complex traffic scenarios, including large-scale simulations and real-world tests with diverse vehicle platoons and traffic. It will particularly address CARL's response to unexpected events like sensor failures and unpredictable human behavior, aiming to boost the system's resilience and reliability for next-generation CAVs.
\textbf{}




 
\bibliography{bare_jrnl.bib}
\bibliographystyle{IEEEtran}

\vspace{11pt}

\vfill

\end{document}